\pdfoutput=1
\documentclass[10pt, logo, copyright]{nvidiatechreport}

\usepackage{graphicx}%
\usepackage{multirow}%
\usepackage{amsmath,amssymb,amsfonts}%
\usepackage{amsthm}%
\usepackage{mathrsfs}%
\usepackage[title]{appendix}%
\usepackage{xcolor}%
\usepackage{textcomp}%
\usepackage{manyfoot}%
\usepackage{booktabs}%
\usepackage{algorithm}%
\usepackage{algorithmicx}%
\usepackage{algpseudocode}%
\usepackage{listings}%

\usepackage{pgfplots}
\pgfplotsset{compat=1.18}
\usepackage{adjustbox}

\title{Reasoning Visual Language Model for Chest X-Ray Analysis.}

\author[1]{Andriy Myronenko}
\author[1]{Dong Yang}
\author[2]{Baris Turkbey}
\author[3]{Mariam Aboian}
\author[4]{Sena Azamat}
\author[5]{Esra Akcicek}
\author[1]{Hongxu Yin}
\author[1]{Pavlo Molchanov}
\author[1]{Marc Edgar}
\author[1]{Yufan He}
\author[1]{Pengfei Guo}
\author[1]{Yucheng Tang}
\author[1]{Daguang Xu}

\correspondingauthor{amyronenko@nvidia.com}

\medskip

\affil[1]{NVIDIA, Santa Clara, CA} 
\affil[2]{NIH/NCI, Bethesda, MD}
\affil[3]{CHOP/UPenn, Philadelphia, PA}
\affil[4]{Basaksehir Cam and Sakura City Hospital, Istanbul, Turkey}
\affil[5]{Lunenfeld-Tanenbaum Research Institute, Toronto, Canada}

\begin{abstract}
Vision–language models (VLMs) have shown strong promise for medical image analysis, but most remain opaque, offering predictions without the transparent, stepwise reasoning clinicians rely on. We present a framework that brings chain-of-thought (CoT) reasoning to chest X-ray interpretation.
Inspired by reasoning-first training paradigms, our approach is designed to learn how experts reason—not just what they conclude—by aligning intermediate steps with observable image evidence and radiology workflow. Beyond accuracy, the explicit reasoning traces support clinical auditability: they reveal why a conclusion was reached, which alternatives were considered, and where uncertainty remains—enabling quality assurance, error analysis, and safer human–AI collaboration. 

\smallskip
Our model couples high-fidelity visual encoding with a two-stage training recipe: a reasoning-style supervised fine-tuning (SFT) followed by reinforcement learning (RL) that uses verifiable rewards over a list of X-ray abnormalities. The model outputs  reasoning that mirrors radiologists’ systematic though process, uncertainty  and differential diagnosis. In out-of-distribution evaluation, the approach achieves competitive multi-label classification while improving interpretability. In a reader study with expert radiologists, full reasoning traces increased confidence, supported error auditing, and reduced time to finalize reports. We release code and model \textbf{NV-Reason-CXR-3B} to facilitate community progress toward trustworthy, explainable AI in chest radiography and broader medical imaging tasks where reasoning quality is as critical as prediction quality.

\smallskip
\textbf{Github Code:} 
\href{https://github.com/NVIDIA-Medtech/NV-Reason-CXR}{https://github.com/NVIDIA-Medtech/NV-Reason-CXR} 

\smallskip
\textbf{Model:} \href{https://huggingface.co/nvidia/NV-Reason-CXR-3BR}{https://huggingface.co/nvidia/NV-Reason-CXR-3B} 

\end{abstract}

\begin{document}








\maketitle


\section{Introduction}
\label{introduction}

Vision--language models (VLMs) are increasingly competitive on medical imaging tasks, yet many systems remain opaque: they generate labels or reports without revealing how conclusions were reached, limiting auditability, error analysis, and clinical trust~\cite{robustness2025}. Radiological expertise is more than a final label; it couples systematic visual search with differential diagnosis, uncertainty calibration, and synthesis into an actionable impression.

We propose a framework that brings chain-of-thought (CoT) reasoning to chest radiography, producing human-interpretable reasoning steps. Inspired by reasoning-first training of DeepSeek-R1~\cite{deepseek2025}, our approach combines a radiologist-style supervised fine-tuning (SFT) warm start with GRPO reinforcement learning (RL) and verifiable rewards defined over a list of chest X-ray abnormalities~\cite{wei2022chainofthought,deepseek2025,shao2024deepseekmath,irvin2019chexpert}.

In an expert reader study, our AI-assisted reasoning increased confidence, supported targeted error auditing, and reduced time to finalize reports—particularly for abnormal cases. On out-of-distribution (OOD) evaluation using the CheXpert test set (with training on MIMIC-CXR), the model attains competitive multi-label classification while providing faithful rationales~\cite{johnson2019mimiccxrjpg,irvin2019chexpert}.

Our contributions are as follows:
\begin{enumerate}
  \item \textbf{Reasoning for CXR via GRPO.} We adapt GRPO to chest X-ray analysis by designing radiology-specific reward functions over verifiable abnormality sets. This enables training with sparse, verifiable signals—without requiring dense human reasoning annotations.
  \item \textbf{Expert reasoning annotations.} We develop a lightweight system to collect high-quality reasoning traces from expert radiologists, capturing their internal thought processes during chest X-ray interpretation via a web-based interface.
  \item \textbf{Synthetic data for SFT.} We introduce a pipeline to synthesize reasoning-style texts for SFT warm start from a set of expert annotations examples. Specifically, we rewrote \(\sim\)100k MIMIC-CXR reports~\cite{johnson2019mimiccxrjpg} with LLMs to match the tone of expert-provided traces.
  \item \textbf{Open source.} We release the model and training code to facilitate reproducibility and further work on trustworthy, auditable medical AI for chest radiography.
\end{enumerate}

\section{Data collection}

\subsection{Human annotations}

We enlisted several experienced radiologists to annotate their internal reasoning while reading chest X-ray cases. To support this, we developed an internal web platform that makes thought capture as seamless as possible. The platform provides automated voice recording, transcription, error correction, and optional translation into English.

We used OpenAI's whisper-large-v3 model to transcribe audio to text~\cite{radford2023whisper}. The system transcribes in the background; radiologists then review the transcript and edit if necessary. We observed frequent transcription errors, particularly for medical terminology. To mitigate this, we added a second pass using an LLM for domain-aware correction. We instructed the LLM that the text concerns chest X-ray radiology to bias corrections toward appropriate terms. This substantially reduced manual edits by radiologists, who previously had to fix spelling and terminology errors. For this correction stage we used the DeepSeek-V3-0324 LLM model~\cite{deepseekv3}.

Our annotation team supported multilingual collaboration, with radiologists recording in English, Turkish, or Vietnamese. Although all participants possessed proficient English skills, they were encouraged to use their native languages similar to their clinical workflow. For non-English recordings, the platform offered automated translation into English, again incorporating radiology-specific context. We utilized the DeepSeek-V3-0324 LLM~\cite{deepseekv3} for these translations.

To promote consistency across annotations, we provided guidelines recommending a structured order for reviewing anatomical regions. While radiologists often develop individualized approaches (e.g., starting with soft tissues or lung fields), this guidance aimed to standardize the ground-truth data. The suggested sequence was as follows:

\begin{enumerate}
  \item Image quality, initial observations
  \item Medical devices, catheters, ECG leads
  \item Central airways (trachea, main bronchi)
  \item Right lung: upper, middle, lower zones
  \item Left lung: upper, lower zones
  \item Mediastinum and hila
  \item Cardiac silhouette/heart region
  \item Upper abdomen
  \item Bones
  \item Summary: major findings and differentials
\end{enumerate}

We further instructed annotators to verbalize their internal thoughts as comprehensively as possible. Unlike standard X-ray reports, which are concise summaries typically spanning one page, we emphasized capturing all elements of reasoning, including hypotheses, visual observations, deliberations, uncertainties, and differential diagnoses. Explicit statements of uncertainty (e.g., unsure or uncertain) were encouraged to reflect authentic cognitive processes.

We asked radiologists to sub-divide each case into short records corresponding to individual thoughts. Generally, each short record captured a distinct idea and/or a distinct anatomical region (e.g., right upper lung). Each record consisted of a paragraph (approximately 1--5 sentences).

\subsubsection{Annotation workflow.}

For each case, the annotation workflow is following:
\begin{enumerate}
  \item The radiologist starts a voice recording. The system automatically transcribes to text (auto-corrects and translates into English when necessary).
  \item The annotator edits the text, verifies correctness, adds missing details, or deletes the record if needed. Sub-dividing into short records allows selective deletion and re-recording without affecting other segments.
  \item The process repeats for subsequent recordings and transcripts. Each record could optionally include a \emph{bounding box}. We provided tools to draw a bounding box over the X-ray to indicate the relevant region. For general statements (e.g., image quality or the final summary), no bounding box was required. For localized statements (e.g., heart region or a focal lesion), a bounding box was encouraged. We also captured mouse-pointer position during the recording. These spatial signals (bounding boxes and pointer movements) were \emph{not} used for current model training and are reserved for future work.
  \item The final record is generally a summary enumerating the most significant findings and potential differential diagnoses.
\end{enumerate}

On average, each full annotation comprised 10-20 records, totaling 7-12 minutes of audio and yielding approximately 1--2 pages of text representing the radiologist's raw chain of thought.

\subsubsection{Image cohort and scale}

We selected a subset of frontal chest X-ray images from~\textbf{ChestX-ray14} due to its permissive license; all voice and text annotations correspond to these images~\cite{wang2017chestxray8}. We selected a diverse cohort of cases, including normal and abnormal studies. Data collection is ongoing; at the time of this model training (September 2025), we collected 100 human-annotated cases, roughly half from US based radiologists and half from international collaborators. An example of an annotated case text is following:

\medskip
\colorbox{gray!10}{
\footnotesize 
\parbox{\dimexpr\textwidth-15\fboxsep\relax}{
\textbf{An example of radiologist annotation text.}

\medskip
We are looking at a PA chest X-ray. The quality of the study is fine. I see that the lung fields are not obscured by any additional structure.

\smallskip
First, we will look for any devices. I see some sternal wires. Here you can see them as radiopacities. The patient probably had some major cardiac or cardiovascular surgery.

\smallskip
We are now looking at the central air pathways. I see that the trachea is patent. Here we are seeing the indentation of the ascending aorta and the aortic arch on the distal third of the trachea. Additionally, both the right and left main bronchi appear patent.

\smallskip
Next, we will evaluate the right upper lobe. The aeration of the right upper lobe is relatively normal. There is no consolidation, and there is no infiltration. However, I see some pleural irregularities at the apex on the right. Again, no evidence of a pneumothorax.

\smallskip
Next, we will look at the right middle and right lower lobes. The aeration is normal. There is no infiltrate. There is no nodule or mass or evidence of pneumothorax.

\smallskip
Now we are looking at the left upper lobe. The left upper lobe appears normal. There is no evidence of a nodule, mass, or pneumothorax.

\smallskip
Now we are looking at the left lower lobe. Again, the visibility is limited. However, I do not see any nodule, mass, or infiltrate. I think there is some minor pleural effusion here. The costophrenic angle is obliterated on the left.

\smallskip
The right hilum lesion shows some prominence of the lung vasculature, primarily the central vasculature.

\smallskip
The left hilum is not well visualized because the cardiothoracic ratio of the patient increased significantly.

\smallskip
Now we are looking at the heart and obviously the cardiothoracic ratio has increased. The heart is enlarged. There is cardiomegaly in this patient. In addition to that, there may be some pericardial effusion contributing to the increased cardiothoracic ratio. It is not easy to confirm from the chest X-ray, but this should also be considered.

\smallskip
Now we are looking at the superior and middle mediastinum. As I mentioned, the aortic arch is prominent and I see some calcifications which are quite significant for underlying hypertension.

\smallskip
We're evaluating the pleura on the right and the costophrenic and cardiophrenic angles are open. There is no evidence of a pleural effusion on the right.

\smallskip
As I mentioned before, on the left, the costophrenic angle is slightly obliterated. This is a significant finding for left-sided minor pleural effusion.

\smallskip
Now we are looking at the imaged portions of the upper abdomen. You are seeing the gastric air. In addition to that, I see some small and large bowel air, which is normal. I don't see any free air or free subdiaphragmatic air.

\smallskip
The final evaluation will be for the bones. I do not see any evidence of a focal lesion or a fracture line in the imaged bones. As I mentioned before, I see the sternotomy wires. The patient likely had cardiothoracic surgery.

\smallskip
Overall, to recap, this patient has significant cardiomegaly and shows signs of hypertension along with minor pleural effusion. This may be related to cardiac failure or chronic cardiac insufficiency.

  }
}

\medskip
Although each human annotation is high quality, we needed more variability and data scale. To this end, we developed an approach to rewrite existing hospital reports in the style of our human reasoning annotations, thereby synthesizing additional reasoning-style data, which is described in the following section.

\subsection{Synthetic data}

Medical labeled data is scarce in general. It is even more pronounced in our scenario, where human internal thinking annotations  are virtually nonexistent.  We therefore designed a system to take advantage of our small set of high quality human labels to guide large-scale generation.

We used X-ray report texts from MIMIC-CXR~\cite{johnson2019mimiccxr, johnson2019mimiccxrjpg} and instructed an LLM to rewrite each report to follow the style of our human reasoning annotations while remaining factual to the underlying report content. We used the GPT-OSS-120b model for rewriting~\cite{openai2025gptoss}.  Its large context window (up to 128k tokens) allowed us to include the human examples and detailed prompts alongside each source report. We prioritized MIMIC-CXR studies with more descriptive reports, as extremely short originals are difficult to expand meaningfully. A total of 100k reports were processed to create a synthetic reasoning dataset. 

\medskip
\colorbox{gray!10}{
\footnotesize 
\parbox{\dimexpr\textwidth-15\fboxsep\relax}{
    \textbf{Prompt template to re-write reports.}
   \begin{itemize}
   \footnotesize 
       \item Re-write the following text so that it closely mimics the characteristics, language choices, implications of the Examples.
       \item Use the same systematic, stepwise approach seen in the examples: view (PA or AP), quality assessment, medical devices, central airways, right lung (upper, middle, and lower lobes) and left lung (upper and lower lobes), pleura, mediastinum, hila, heart, upper abdomen, and bony structures.
        \item Do not mention lateral views if present in the input report.
        \item Comment both on the presence and absence of medical devices and abnormalities.
        \item Use first-person narrative.
        \item Use differential diagnoses blended into observations.
        \item Direct address to the reader as in examples.
        \item Remove all references to prior exams, converting any comparative language of prior exams into \item End with a concise summary of the most significant findings and their possible clinical implications, without recommendations or comparisons to previous studies.
        \item Output only the rewritten report, without any additional text, intro, titles, section headers, lists or bullet points.
   \end{itemize}
  }
}

\medskip
Conventional chest X-ray reports are typically concise (approximately 1--2 paragraphs). The rewritten versions expand to approximately 1--2 pages, adding structured narrative, explicit hypotheses, and rationale. While the LLM attempts to remain factual to the source report, added verbosity can introduce \emph{noise} (e.g., mild speculation). Our premise is that diversity and volume help the model learn robust reasoning patterns despite this noise. We used these 100k synthetic cases for the first training stage of our VLM.

An example of an input X-ray report and the rewritten synthetic data is following:

\medskip
\colorbox{gray!10}{
\footnotesize 
    \parbox{\dimexpr\textwidth-15\fboxsep\relax}{
    \textbf{Input chest X-ray report text.}
    
\medskip
 EXAMINATION:  CHEST (PA AND LAT)
 
\smallskip
 INDICATION:  Evaluate right-sided chest tube

 \smallskip
 FINDINGS: 
 There is a small right-sided pneumothorax.  A chest tube is noted at the right
 lung base.  There is a small right-sided pleural effusion with associated
 parenchymal opacity which could reflect atelectasis. The heart is enlarged.
 The visualized left hemithorax is clear.

  }
}

\medskip
\colorbox{gray!10}{
\footnotesize 
    \parbox{\dimexpr\textwidth-15\fboxsep\relax}{
    \textbf{Synthetically rewritten report.}
    
\medskip
We'll begin with the quality assessment of this PA chest X-ray. The quality is acceptable, and the lung fields are well visualized, allowing for a thorough evaluation. As you can see, there are no significant obstructions to the lung parenchyma.

\smallskip
Now, we are looking for medical devices or instrumentation. I do see a chest tube, which is positioned at the right lung base, suggesting a prior procedure to address a condition affecting the right lung. 

\smallskip
We're checking the central airways, and the trachea appears patent. The right and left main bronchi also appear patent, which is a good sign. 

\smallskip
Moving on to the right lung, we'll evaluate the upper lobe first. The aeration of the right upper lobe appears relatively normal. However, when we look at the lower lobe, we notice a small pneumothorax. This is a significant finding, as it indicates that there is air in the pleural space, which can be a serious condition. Additionally, there is a small right-sided pleural effusion associated with parenchymal opacity, which could reflect atelectasis. This suggests that there might be some fluid accumulation in the pleural space, possibly due to the pneumothorax or an underlying infection.

\smallskip
Now, let's examine the left lung. The left upper and lower lobes appear clear, with no evidence of infiltrates, masses, or pneumothorax. This is a positive sign, indicating that the left lung is not significantly affected by the condition that required the chest tube.

\smallskip
We'll proceed to evaluate the pleura on both sides. On the right, we've already noted a small pleural effusion. On the left, the costophrenic and cardiophrenic angles appear normal, with no evidence of pleural effusion.

\smallskip
Next, we'll assess the mediastinum. The heart size appears enlarged, which could be meaningful for cardiomegaly. This is an important finding, as it may indicate an underlying cardiac condition.

\smallskip
The hila of both lungs appear unremarkable, with no significant adenopathy or masses. 

\smallskip
We're also evaluating the upper abdomen, although it's not fully visualized. We don't see any free subdiaphragmatic air, which is a good sign.

\smallskip
Finally, we are looking at the bony structures. The clavicles, scapula, and ribs appear normal, with no focal lesions or fractures. 

\smallskip
To summarize our findings, the most significant findings in this patient are the small right-sided pneumothorax, the pleural effusion with associated parenchymal opacity, and the enlarged heart. These findings could represent a complex condition involving the right lung and possibly the heart, with the pneumothorax and pleural effusion suggesting a need for close monitoring and possibly further intervention.
  }
}
\medskip

\subsection{Refusal and Hallucination Controls}

To improve practicality and reduce unintended behavior, we interleaved a small refusal set during training, with a default fallback:
\textit{``I'm sorry, I'm specialized in medical chest X-ray analysis.''}
The refusal data covered:
\begin{itemize}
    \item Natural (non-medical) images.
    \item Medical but non–X-ray images (e.g., ultrasound, CT, MRI 2D slices).
    \item Irrelevant follow-ups (trivia such as ``What is the capital of France?'', random keyboard input, etc.). A subset of such prompts was generated with GPT-OSS-120B.
\end{itemize}
This set is not intended to be comprehensive, but it reduces responses to accidental inputs (non-medical images, off-topic text) and decreases off-distribution hallucinations. 

At inference time, additional guardrails can be layered without retraining~\cite{nemo_guardrails}.

\subsection{Multi-Turn Follow-Up Chat Data}

It is common for AI systems to receive follow-up questions (e.g., ``Where do you see the lesion?'' or ``Provide a differential diagnosis.''). To handle such interactions, we generated follow-up question–answer (QA) pairs from a synthetic dataset using the GPT-OSS-120B LLM. Prompts covered a broad range of inquiry types—\emph{what, where, why, how, list, describe, explain, identify, compare, classify, quantify, evaluate, predict, summarize, differentiate, interpret, suggest}, etc.—to probe understanding of findings. Each answer included a short ``thinking'' paragraph first, followed by the concise reply. In total, we produced \(\sim\)500K QA pairs.

\subsection{Structured Report Query}

A special type of follow-up request is ``Write a structured report,'' an essential component of the radiology workflow. For a new chest X-ray, a clinical radiologist typically reviews the study and submits a structured report; the exact format varies across institutions. We follow the template outlined in~\cite{marcovici2014structured} to generate structured reports for all cases. Full Reasoning traces are rewritten into this structured form using the GPT-OSS-120B LLM.  An example of the structured report is following:

\medskip
\colorbox{gray!10}{
\footnotesize 
    \parbox{\dimexpr\textwidth-15\fboxsep\relax}{
    
\medskip
INDICATION:
Not provided.

\smallskip
COMPARISON:
None.

\smallskip
TECHNIQUE:
PA view of the chest.

\smallskip
FINDINGS:
Medical devices: Chest tube at right lung base.
Central airways: Trachea and main bronchi patent.
Lungs and pleural space: Small right pneumothorax; small right pleural effusion; parenchymal opacity in right lung; left lung clear.
Mediastinum and hila: Mediastinum normal; hila normal; no adenopathy.
Heart: Enlarged heart.
Upper abdomen and diaphragm: No free subdiaphragmatic air; diaphragm normal.
Bones and soft tissues: Normal.

\smallskip
IMPRESSION:
\begin{itemize}
\footnotesize 
    \item Small right-sided pneumothorax.
    \item Small right pleural effusion with parenchymal opacity.
    \item  Enlarged heart    
\end{itemize}

  }
}
\medskip

Producing a structured-report follow-up enables streamlined clinical use: the output can be copied into reporting systems (with minimal manual edits), improving efficiency, saves time and encouraging completeness/consistency~\cite{marcovici2014structured}.

\section{Methods}
\label{methods}

\subsection{Overview}
Recent progress in open reasoning models—most notably DeepSeek-R1~\cite{deepseek2025} - shows that one can train for reasoning quality using reinforcement learning (RL) with verifiable answers, avoiding the need for dense, token-level supervision. While R1 was originally designed for math, logic, and code, its training paradigm is well suited to medical imaging, where granular annotations (e.g., full chain-of-thought traces) are scarce and expensive. We adopt this paradigm for chest radiography by (1) warm-starting the model with a set of radiologist-style reasoning demonstrations via supervised fine-tuning (SFT), then (2) continuing training with Group Relative Policy Optimization (GRPO)~\cite{shao2024deepseekmath} using rewards derived from abnormality labels.

Our system comprises two stages:
\begin{enumerate}
  \item Supervised Finetunning (SFT)  to follow radiologist-like reasoning format and vocabulary using high-quality reasoning traces.
  \item GRPO reinforcement learning to continue training on larger dataset and prefer trajectories whose generated rationale and final impression — match verifiable abnormalities, without requiring ground-truth chain-of-thought.
\end{enumerate}

\subsection{Problem Setup}
Given a chest X-ray $x$ and associated verifiable supervision $z$, our model $\pi_\theta(y \mid x)$ produces a free-form rationale followed by a structured impression. Let $\mathcal{A}$ denote a discrete ontology of findings (e.g., \texttt{cardiomegaly}, \texttt{pleural\_effusion}, \texttt{consolidation}, \dots). For each training case we have image-level labels $z_{\text{cls}} \subseteq \mathcal{A}$ (multi-label abnormalities). We define a deterministic parser $\Phi(\cdot)$ that extracts a predicted abnormality set $\hat{z}_{\text{cls}}$ from the model’s output $y$ using pattern rules. We format each model output as
\verb|<think> ... </think> <answer> ... </answer>|
so that the final answer can be deterministically parsed from the
\verb|<answer>...</answer>| span.

\subsection{Stage I: SFT}

We use traditional finetuning by minimizing standard next-token loss over a curated set of $(x, y^{\star})$ pairs where $y^{\star}$ contains stepwise, radiologist-style reasoning:
\[
\mathcal{L}_{\text{SFT}}(\theta) = - \mathbb{E}_{(x, y^\star)} \sum_{t=1}^{T} \log \pi_\theta\!\left(y^\star_t \mid x, y^\star_{<t}\right).
\]
This teaches format, tone, and clinically appropriate decomposition, providing a stable initialization for RL.

\subsection{Stage II: Group Relative Policy Optimization (GRPO)}
We then switch to GRPO~\cite{shao2024deepseekmath}, an RL algorithm that replaces value-function learning with groupwise relative advantages computed from simple, verifiable rewards.

\paragraph{Sampling and scoring.}
To briefly recap the GRPO, for each prompt $x$, we sample a group of $G$ completions $y_{1:G} \sim \pi_\theta(\cdot \mid x)$ and compute a scalar reward $r_i = R(x, y_i, z)$ per completion. Rewards are normalized within the group:
\[
\tilde{r}_i = \frac{r_i - \mu_r}{\sigma_r + \epsilon}, \quad
\mu_r = \frac{1}{G}\sum_{j=1}^G r_j, \quad
\sigma_r = \sqrt{\frac{1}{G}\sum_{j=1}^G (r_j - \mu_r)^2}.
\]

The policy update maximizes the likelihood of higher-scoring samples while suppressing lower-scoring ones, with a KL control to the SFT policy $\pi_{\theta_0}$:
\[
\mathcal{L}_{\text{GRPO}}(\theta) = - \mathbb{E}_{x}\,\frac{1}{G}\sum_{i=1}^{G} 
\tilde{r}_i \sum_{t=1}^{T_i} \log \pi_\theta(y_{i,t} \mid x, y_{i,<t})
\;+\; \beta \, \mathrm{KL}\!\left(\pi_\theta(\cdot\mid x)\,\Vert\,\pi_{\theta_0}(\cdot\mid x)\right).
\]

Following recent GRPO variants, we set $\beta=0$, removing the KL anchor to the SFT reference, which can be beneficial when the reference is misaligned with the downstream reward or when fast adaptation is needed.  Preference-optimization methods show that strong performance is possible even without an explicit KL, relying instead on implicit regularization~\cite{rafailov2023direct}.
In our setting, the SFT initializer is trained on noisy synthetic traces, and our objective prioritizes correctness of abnormality predictions over stylistic conformity; dropping the KL thus enables larger reward-aligned policy moves while avoiding over-anchoring to suboptimal language patterns.

\subsection{Verifiable Rewards for Chest X-Rays}
The advantage of rewards is they require no chain-of-thought text; they depend only on discrete list of abnormalities.   In the original GRPO formulation the rewards are binary and based on the correctness of the answer.  In our case we predict a list of abnormalities, not a single answer, so we introduce a novel X-ray specific reward functions. 

\paragraph{(1) Abnormality set correctness reward.}
 The main reward function we use is \emph{abnormality set correctness}, which is an intersection over union metric between predicted and ground-truth sets.
Let $\hat{z}_{\text{cls}}=\Phi_{\text{cls}}(y)$ be the set extracted from $y$. 

The image–level supervision set \(z_{\text{cls}}\) follows the CheXpert
label ontology~\cite{irvin2019chexpert}. Specifically, we use
\(\mathcal{A}=\{\)\texttt{Atelectasis}, \texttt{Cardiomegaly},
\texttt{Consolidation}, \texttt{Edema}, \texttt{Enlarged Cardiomediastinum},
\texttt{Fracture}, \texttt{Lung Lesion}, \texttt{Lung Opacity},
\texttt{No Finding}, \texttt{Pleural Effusion}, \texttt{Pleural Other},
\texttt{Pneumonia}, \texttt{Pneumothorax}, \texttt{Support Devices}\(\}\).

We define the abnormality set correctness reward with optional label weighting $w_a$:
\[
r_{\text{cor}}(y) = 
\frac{\sum_{a \in \mathcal{A}} w_a \cdot \mathbf{1}[a \in \hat{z}_{\text{cls}} \cap z_{\text{cls}}]}
{\sum_{a \in \mathcal{A}} w_a \cdot \mathbf{1}[a \in \hat{z}_{\text{cls}} \cup z_{\text{cls}}]}.
\]
Weights can upweight safety-critical findings (e.g., Lung Lesion), but in practice we used equal weights $w_a==1$. This reward function is by definition bound to (0, 1) interval.

\paragraph{(2) Format reward}
To ensure the learned response preserves the required structure, we use a binary format reward \(r_{\text{fmt}}\in\{0,1\}\). Concretely, \(r_{\text{fmt}}=1\) iff all tags \verb|<think>|, \verb|</think>|, \verb|<answer>|, \verb|</answer>| are present \emph{exactly once}; otherwise \(r_{\text{fmt}}=0\). We verify this with deterministic text parsing prior to scoring.

\paragraph{(3) Overshort penalty}

We found it beneficial to introduce a \emph{length-aware} reward that discourages overly short responses. While many RL setups implicitly favor short responses (e.g., via per-token log-prob objectives or explicit length penalties on long generations), our setting requires the model to articulate a systematic, radiologist-style search pattern even when the abnormality set is trivial (e.g., \textit{No Finding}). Without an explicit constraint, the policy quickly learns to truncate its reasoning once the abnormality list is correct, skipping normal-but-necessary observations (airways, bones, cardiac silhouette, lungs, devices). Empirically, useful rationales fall in the 600--700 token range; we therefore apply a soft penalty to completions shorter than $L_{\min}=400$ tokens.

Let $L(y)$ denote the token length of completion $y$. We define a bounded, \textbf{overshort penalty}
\[
r_{\text{len}}(y) = \min\left(0,\ \frac{L(y) - L_{\min}}{L_{\min}}\right)
\in [-1..0],
\]
which is $0$ when $L(y) == L_{\min}$ and decreases linearly up to $-1$ as $L(y)\to 0$.

\paragraph{Composite reward.}
The final composite reward is defined as :
\[
r = r_{\text{cor}} * r_{\text{fmt}} +  r_{\text{len}}
\]
Here, the binary format reward is used as a hard threshold (instead of addition) - if the format is wrong - the correctness reward is set to 0. We found it empirically to work better to prevent accidental incorrect formatting.

\subsection{Training Details}

\textbf{Backbone.}
We adopt \textbf{Qwen2.5-VL-3B-Instruct}\footnote{https://hf.co/Qwen/Qwen2.5-VL-3B-Instruct}, a vision–language model that combines a ViT-based vision encoder with a decoder-only LLM. 
Qwen2.5-VL features several architectural and training improvements, e.g. an efficient vision stack (windowed attention in ViT) and alignment to the language backbone with RMSNorm and SwiGLU—which together yield strong zero-/few-shot multimodal performance while remaining parameter-efficient for domain adaptation~\cite{bai2025qwen25vl}.

\textbf{Input preprocessing.}
Unless otherwise noted, images are resized proportionally to less than 476x476 pixels total.

\subsection{SFT}
We initialize from the instruction-tuned checkpoint and perform a SFT on a synthetic dataset, which included 100K synthetic reasoning traces and 500K follow-up multi turn QA pairs.  This stage serves multiple purposes: precondition image encoder for X-ray images, teaches formatting and image-to-observation correctness, providing a stable starting point for subsequent RL.

The training was performed on 4 nodes of 8xH100 NVIDIA GPU machines. All experiments use $bfloat16$  training with gradient check-pointing, gradient accumulations steps of 8. The batch size was 1 per gpu, optimized with AdamW optimizer, with cosine scheduler starting with a learning rate of $2e^{-5}$. Grad norm clipping was set to 0.3. The optimization ran for 2 epochs.

\subsection{RL with GRPO.}

The training was performed on 4 nodes of 8xH100 NVIDIA GPU machines. All experiments use $bfloat16$  training with gradient checkpointing, gradient accumulation steps of 8.
For GRPO each prompt has a number of generations equal to 16, with temperature of 1. Each gpu was set to handle 1 prompt.  AdamW optimizer was used with cosine schedule starting from a learning rate of $1e^{-6}$.  Grad norm clipping was set to 0.2. Optimization was run for 10 epochs.

\subsection{Implementation details}

Many design choices were inspired by Open-R1\footnote{https://github.com/huggingface/open-r1} - a great open source project by Hugginface aiming to reproduce DeepSeek-R1~\cite{deepseek2025} performance for text only inputs in PyTorch.  Both SFT and GRPO training loops take advantage of several Hugginface libraries, including TRL, Transformers and Accelerate. We make our training code public\footnote{https://github.com/NVIDIA-Medtech/NV-Reason-CXR}.

\section{Results}
\label{results}

\subsection{Subjective Evaluation with Expert Radiologists}

We conducted a within-subject user study with US board-certified radiologists to assess perceived quality, usefulness, time savings and safety of the model’s outputs. Each reader interpreted the same chest X-ray cases under three assistance conditions:
\begin{enumerate}
    \item \textbf{Manual Baseline:} no AI assistance;
    \item \textbf{Labels only:} AI-provided list of predicted abnormalities (without any reasoning text);
    \item \textbf{Full AI reasoning:} Full AI reasoning output and the structured report.
\end{enumerate}

Readers were instructed to behave as in routine practice.

\paragraph{Participants and Materials.}
\begin{itemize}
    \item \textbf{Participants:} 2 radiologists (10+ years-in-practice recorded).
    \item \textbf{Cases:} 10 cases (2 normal, and 8 abnormal).
    \item \textbf{Platform:} web-interface of the trained AI model\footnote{https://huggingface.co/spaces/nvidia/nv-reason-cxr}
\end{itemize}

\subsubsection{Procedure.}
For each case:
\begin{enumerate}
    \item Reader interprets without AI following their clinical routine. Specifically, the radiologist voice dictated the chest X-ray report (using Nuance~\cite{nuance_powerscribe_one} and manually corrected, including all hospital required section, section Findings, Impressions, etc. The total time spent is recorded (e.g. 4 min)
    \item After a short distraction, the case is re-presented with the \emph{abnormalities-only} list (AI generated); the reader may revise findings/impressions and time.
    \item After a second distraction, the case is shown with the \emph{full reasoning} text and its structured report form.  The reader can copy-paste the structured report, double check it and edit if necessary. The total time is recorded again.
    \item After each condition, the reader completes the Likert questionnaire; free-text comments were encouraged.
\end{enumerate}
At each step, only information permitted for that condition is visible.

\subsubsection{Questionnaire Items.}

All items use a 5-point \textbf{Likert Scale}: \emph{1 = Strongly disagree, 2 = Disagree, 3 = Neither agree nor disagree, 4 = Agree, 5 = Strongly agree}.

\paragraph{A. Accuracy and Reasoning Quality}
\begin{itemize}
    \item The reasoning trace includes correct statements about key observations.
    \item The reasoning does not include speculative or unsupported statements.
    \item The AI identified a relevant finding I initially missed.
    \item Uncertainty and limitations are explicitly stated when appropriate.
\end{itemize}

\paragraph{B. Time \& Efficiency}
\begin{itemize}
    \item The AI output reduced the time required to compose my report.
    \item The AI output helped me reach a decision faster.
    \item Approximate percentage of time saved on this case (percentage calculated)
\end{itemize}

\paragraph{C. Trust and Confidence}
\begin{itemize}
    \item After reading the AI output, I feel more confident (less uncertain) in my decision.
    \item If the AI had disagreed with me on this case, I would re-evaluate and re-examine the study.
    \item The AI’s suggestions improved my final decision.
    \item The AI did not bias me toward incorrect conclusions. 
\end{itemize}

\subsubsection{Reporting}

\definecolor{NVGreenColor}{RGB}{118,185,0} 
\definecolor{CadmiumColor}{RGB}{210,43,43}   

\begin{figure*}[t]
\begin{tikzpicture}
\begin{axis}[
  xbar,
  width=0.85\linewidth,
  height=0.35\textheight,          
  xmin=1, xmax=5.5,                         
  xlabel={Likert score (1--5)},
  ytick={1,2,3,4},
  y dir=reverse,                           
  yticklabels={
    {The reasoning trace includes correct statements about key observations.},
    {The reasoning does not include speculative or unsupported statements.},
    { The AI identified a relevant finding I initially missed.},
    {Uncertainty and limitations are explicitly stated when appropriate.},
  },
  yticklabel style={text width=.22\linewidth, align=right, font=\footnotesize},
  enlarge y limits=0.1,
  bar width=10pt,
  xmajorgrids=true,
  legend style={at={(0.5,1.02)}, anchor=south, legend columns=-1, font=\footnotesize},
   ytick pos=left, 
]

\addplot+[area legend, fill=NVGreenColor, error bars/.cd, x dir=both, x explicit] coordinates {
  (4.5,1) +- (0.54,0)
  (4.67,2) +- (0.51,0)
  (3.83,3) +- (1.17,0)
  (4.67,4) +- (0.51,0)

};
\legend{Full Reasoning (AI-assisted)}
\end{axis}
\end{tikzpicture}%

\caption{The results (mean/std) of Accuracy and Reasoning Quality survey. Expert radiologists were tasked to write a report given the AI full reasoning output. Likert scale: 1 - Strongly Disagree, 2 - Disagree, 3 - Neither Agree Nor Disagree, 4 - Agree, 5 - Strongly Agree. The AI-assisted results demonstrated high average Likert scores in the Accuracy and Reasoning Quality evaluation.}
\label{fig:results_survey_acc}
\end{figure*}
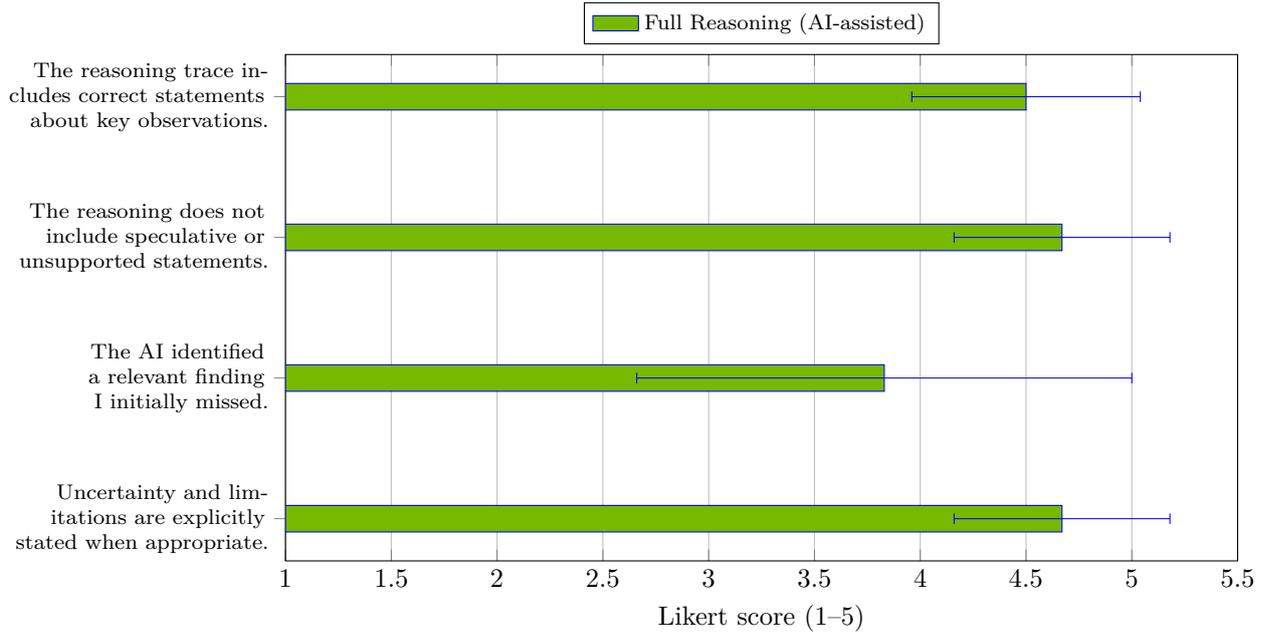

\paragraph{Accuracy and Reasoning Quality.}
Accuracy and Reasoning Quality survey evaluation were conducted for AI-assisted results with full reasoning text provided (see Figure~\ref{fig:results_survey_acc}). The AI model scored very favorably in this task with  a high Likert scores:  
\begin{itemize}
    \item \textbf{The reasoning trace includes correct statements about key observations}. The reasoning traces generally contained accurate statements about salient findings; the majority of responses were \emph{Agree} or \emph{Strongly Agree}, with a mean Likert score of $\sim\!4.5$.
    \item \textbf{The reasoning does not include speculative or unsupported statements}. Most readers selected \emph{Strongly Agree} (mean $\sim\!4.7$). Lower ratings typically reflected specific errors, e.g., an erroneously reported “tube” in one case.
    \item \textbf{The AI identified a relevant finding I initially missed}. Ratings ranged from \emph{Neither} to \emph{Strongly Agree} (mean $\sim\!3.8$). Because participants were expert radiologists, their initial conclusions often matched the AI’s. In some cases, however, the AI added useful details—for example, flagging a \emph{tracheal deviation to the right} that a reader had overlooked.
    \item \textbf{Uncertainty and limitations are explicitly stated when appropriate}.  Most responses were \emph{Agree} or \emph{Strongly Agree} (mean $\sim\!4.6$). Readers highlighted appropriate use of uncertainty language (e.g., listing plausible differentials or explicitly noting uncertainty in challenging cases).
\end{itemize}

\paragraph{Time \& Efficiency.}

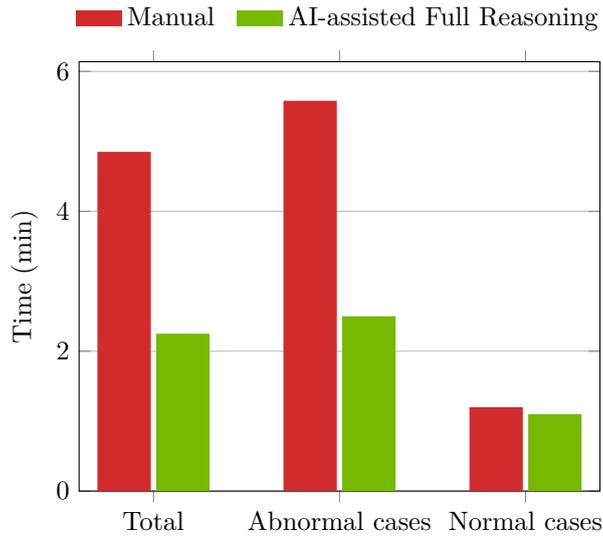
\begin{figure*}[t]
\centering
\begin{tikzpicture}
\begin{axis}[
  ybar,
  bar width=20pt,
  ylabel={Time (min)},
  symbolic x coords={Total, Abnormal cases, Normal cases},
  xtick=data,
  ymin=0,
  enlarge x limits=0.2,
  ymajorgrids=true,
  nodes near coords style={font=\scriptsize, anchor=south},
  nodes near coords align={vertical},
  legend style={at={(0.5,1.05)}, anchor=south, legend columns=-1, /tikz/every even column/.append style={column sep=6pt}, draw=none},
]
\addplot+[area legend, fill=CadmiumColor, draw opacity=0] coordinates {(Total, 4.85) (Abnormal cases, 5.58) (Normal cases, 1.2)};
\addplot+[area legend, fill=NVGreenColor, draw opacity=0] coordinates {(Total, 2.25) (Abnormal cases, 2.5) (Normal cases, 1.1)};

\legend{Manual, AI-assisted Full Reasoning}
\end{axis}
\end{tikzpicture}
\caption{Average time spent interpreting a chest X-ray and submitting a report. When the AI’s reasoning text and a pre-populated structured report were provided, we observed substantial time savings—especially for abnormal cases.}
\label{fig:results_time}
\end{figure*}

We first compare the time \emph{spent} on manual report creation versus AI-assisted reporting in Figure~\ref{fig:results_time}. For the manual condition, reports were voice-dictated using Nuance software~\cite{nuance_powerscribe_one} and pre-populated with standard section headers (FINDINGS, IMPRESSION, etc.) to simulate a realistic clinical workflow. For the AI-assisted condition with full reasoning output, the model’s structured report was also provided so that readers could copy–paste and then edit as needed (via dictation or typing).

For \textbf{abnormal cases}, report submission was nearly $2\times$ faster with AI assistance (Figure~\ref{fig:results_time}): average read time $\sim\!5.5$\,min manually versus $\sim\!2.5$\,min with AI. Although radiologists still confirmed all findings and adjusted the text, AI assistance yielded substantial time savings. For \textbf{normal cases}, timings were nearly identical across conditions, as modern hospital systems typically start with a normal-case template requiring minimal edits.

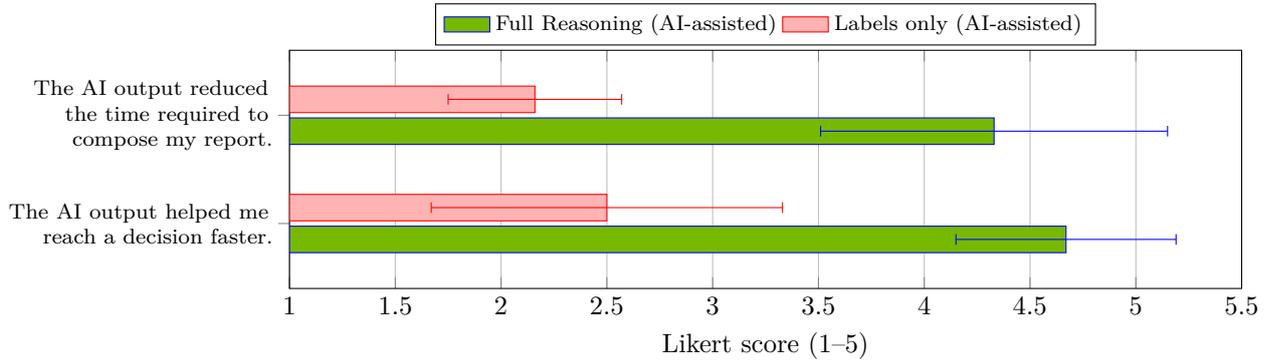
\begin{figure*}
\begin{tikzpicture}
\begin{axis}[
  xbar,
  width=0.85\linewidth,
  height=0.2\textheight,          
  xmin=1, xmax=5.5,                         
  xlabel={Likert score (1--5)},
  ytick={1,2},
  y dir=reverse,                           
  yticklabels={
    {The AI output reduced the time required to compose my report.},
    {The AI output helped me reach a decision faster.}
  },
  yticklabel style={text width=.22\linewidth, align=right, font=\footnotesize},
  enlarge y limits=0.6,
  bar width=10pt,
  xmajorgrids=true,
  legend style={at={(0.5,1.02)}, anchor=south, legend columns=-1, font=\footnotesize},
   ytick pos=left, 
]

\addplot+[area legend, fill=NVGreenColor, error bars/.cd, x dir=both, x explicit] coordinates {
  (4.33,1) +- (0.82,0)
  (4.67,2) +- (0.52,0)
};

\addplot+[area legend,  error bars/.cd, x dir=both, x explicit] coordinates {
  (2.16,1) +- (0.41,0)
  (2.5,2) +- (0.83,0)
};

\legend{Full Reasoning (AI-assisted), Labels only (AI-assisted)}
\end{axis}
\end{tikzpicture}%

\caption{The Time \& Efficiency survey (mean/std) was evaluated under two AI-assisted scenarios: (i) Full Reasoning (complete rationale plus findings) and (ii) Labels-only (findings list without explanatory text). Likert scale: 1 - Strongly Disagree, 2 - Disagree, 3 - Neither Agree Nor Disagree, 4 - Agree, 5 - Strongly Agree. In nearly all items, the Full Reasoning condition achieved high Likert scores, indicating substantial time savings when writing chest X-ray reports. By contrast, the Labels-only condition scored low, with negligible benefit for time savings.}
\label{fig:results_survey_time}
\end{figure*}

The Time \& Efficiency survey was conducted under two AI-assisted scenarios: \emph{Labels-only} and \emph{Full Reasoning}. In the \emph{Labels-only} scenario, radiologists received only a list of predicted abnormalities (e.g., pneumothorax, atelectasis, support devices), mimicking traditional classifier outputs without explanations. In the \emph{Full Reasoning} scenario, radiologists were provided the model’s complete rationale and a structured report (see Figure~\ref{fig:results_survey_time}).  More specifically:

\begin{itemize}
    \item \textbf{The AI output reduced the time required to compose my report.} With \emph{Full Reasoning} output, nearly all cases were rated \emph{Agree} or \emph{Strongly Agree}, consistent with observed time savings. In contrast, under the \emph{Labels-only} condition most cases were rated \emph{Disagree}. Readers noted that a label alone (e.g., \emph{pneumothorax}) still required re-reading the study to locate and describe the finding, yielding negligible time savings.
    \item \textbf{The AI output helped me reach a decision faster.} Similarly, with \emph{Full Reasoning} most readers selected \emph{Agree} or \emph{Strongly Agree}, whereas the \emph{Labels-only} scenario provided little benefit for decision speed.
\end{itemize}

\paragraph{Trust \& Confidence.}

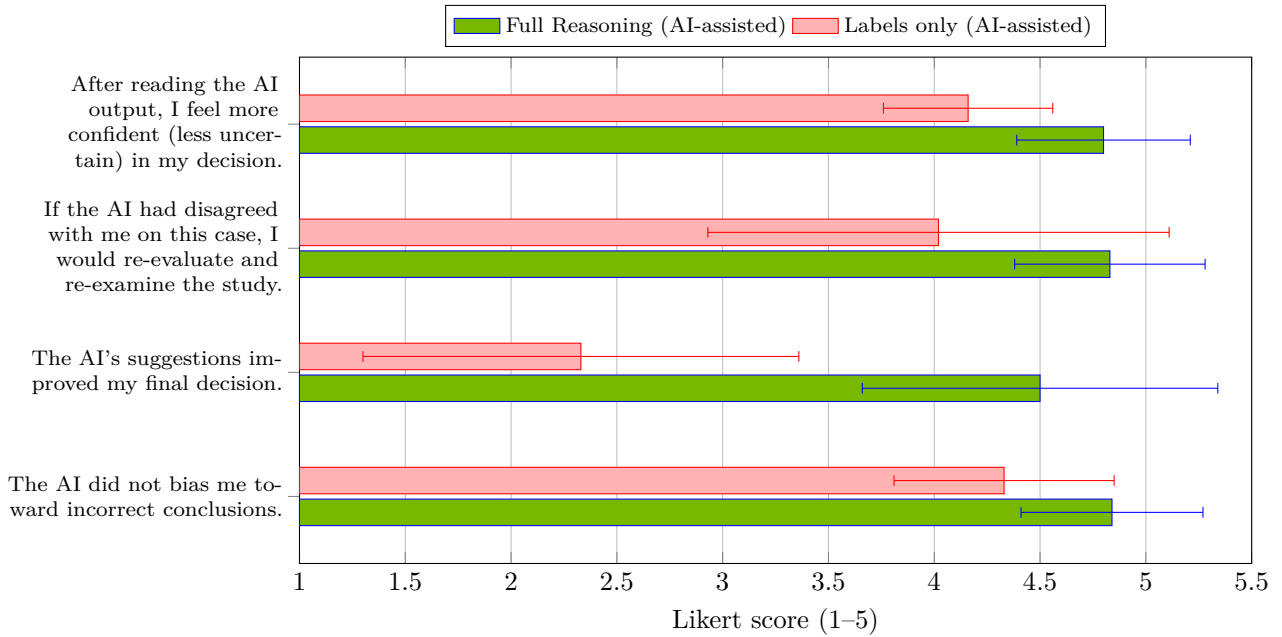
\begin{figure*}
\begin{tikzpicture}
\begin{axis}[
  xbar,
  width=0.85\linewidth,
  height=0.35\textheight,          
  xmin=1, xmax=5.5,                         
  xlabel={Likert score (1--5)},
  ytick={1,2,3,4},
  y dir=reverse,                           
  yticklabels={
    {After reading the AI output, I feel more confident (less uncertain) in my decision.},
    {If the AI had disagreed with me on this case, I would re-evaluate and re-examine
the study.},
    {The AI’s suggestions improved my final decision.},
    {The AI did not bias me toward incorrect conclusions.}
  },
  yticklabel style={text width=.22\linewidth, align=right, font=\footnotesize},
  enlarge y limits=0.18,
  bar width=10pt,
  xmajorgrids=true,
  legend style={at={(0.5,1.02)}, anchor=south, legend columns=-1, font=\footnotesize},
   ytick pos=left, 
]

\addplot+[area legend, fill=NVGreenColor, error bars/.cd, x dir=both, x explicit] coordinates {
  (4.8,1) +- (0.41,0)
  (4.83,2) +- (0.45,0)
  (4.5,3) +- (0.84,0)
  (4.84,4) +- (0.43,0)
};

\addplot+[area legend,  error bars/.cd, x dir=both, x explicit] coordinates {
  (4.16,1) +- (0.40,0)
  (4.02,2) +- (1.09,0)
  (2.33,3) +- (1.03,0)
  (4.33,4) +- (0.52,0)
};

\legend{Full Reasoning (AI-assisted), Labels only (AI-assisted)}
\end{axis}
\end{tikzpicture}%

\caption{The Trust \& Confidence survey (mean/std) was evaluated under two AI-assisted scenarios: (i) Full Reasoning (complete rationale plus findings) and (ii) Labels-only (findings list without explanatory text). Likert scale: 1 - Strongly Disagree, 2 - Disagree, 3 - Neither Agree Nor Disagree, 4 - Agree, 5 - Strongly Agree. In nearly all items, the Full Reasoning condition achieved high Likert scores, indicating high level of trust and benefits of AI-assisted results. }
\label{fig:results_survey_trust}
\end{figure*}

The Trust \& Confidence survey was also conducted under two AI-assisted scenarios: \emph{Labels-only} and \emph{Full Reasoning} (see Figure~\ref{fig:results_survey_trust}). More specifically:

\begin{itemize}
    \item \textbf{After reading the AI output, I feel more confident (less uncertain) in my decision.}
    Across AI-assisted cases, ratings were mostly \emph{Agree} or \emph{Strongly Agree}. The \emph{Full Reasoning} condition scored roughly half a point higher, with more \emph{Strongly Agree} responses, indicating that even when the AI merely confirmed the radiologist’s conclusion, it increased confidence—most notably when full reasoning was provided.

    \item \textbf{If the AI had disagreed with me on this case, I would re-evaluate and re-examine the study.}
    In the \emph{Labels-only} scenario, responses clustered around \emph{Agree} or \emph{Neither}. With \emph{Full Reasoning}, most readers selected \emph{Strongly Agree}, indicating they would re-evaluate. For example, one reader reported rethinking and refining the conclusion after the AI connected cardiomegaly with diffuse opacity and suggested heart failure as a differential. The multi-step reasoning helped integrate findings beyond routine, time-pressured workflows.

    \item \textbf{The AI’s suggestions improved my final decision.}
    We observed a marked advantage for \emph{Full Reasoning} over \emph{Labels-only}. The latter was generally rated \emph{Disagree}, whereas \emph{Full Reasoning} was mostly \emph{Agree}/\emph{Strongly Agree}. In one case, the AI argued the findings were most consistent with a \emph{granuloma}, providing supporting evidence (ultimately confirmed), whereas the manual report was initially vague; the AI output enabled a more precise, defensible impression.

    \item \textbf{The AI did not bias me toward incorrect conclusions.}
    Both AI-assisted scenarios were rated highly (\emph{Agree}/\emph{Strongly Agree}), with \emph{Full Reasoning} scoring higher, likely due to greater transparency behind its recommendations.
\end{itemize}

\subsection{Numeric Evaluation}

We evaluate model performance in terms of multi-label classification accuracy on the final \emph{list of detected abnormalities}. In this setup, only the output in \verb|<answer>| is used to compute standard metrics enabling direct comparison to non-reasoning AI methods. 

Although the full reasoning output is much more than a final label and contains substantially amount information, assigning numeric scores to it for evaluation is non-trivial. We examined text-based metrics (e.g., BLEU~\cite{papineni2002bleu}, ROUGE~\cite{lin2004rouge}), but empirically observed weak correlation with diagnostic correctness, as these measures primarily capture surface-form similarity rather than clinical validity.

For classification evaluation, we follow the MedGemini protocol~\cite{saab2024medgemini_capabilities,yang2024advancing_medgemini}: train on MIMIC-CXR~\cite{johnson2019mimiccxrjpg} and evaluate out-of-distribution (OOD) on the CheXpert test set~\cite{irvin2019chexpert}. 

MIMIC-CXR dataset~\cite{johnson2019mimiccxrjpg} is a large-scale collection of 227,827 chest X-ray studies,
paired with free-text radiology reports.  It includes automatically generated labels for 14 classes (12 abnormalities, Support Devices and No Findings classes) derived via the CheXpert labeler~\cite{irvin2019chexpert}.  The full list of classes include: \(\mathcal{A}=\{\)\texttt{Atelectasis}, \texttt{Cardiomegaly},
\texttt{Consolidation}, \texttt{Edema}, \texttt{Enlarged Cardiomediastinum},
\texttt{Fracture}, \texttt{Lung Lesion}, \texttt{Lung Opacity},
\texttt{No Finding}, \texttt{Pleural Effusion}, \texttt{Pleural Other},
\texttt{Pneumonia}, \texttt{Pneumothorax}, \texttt{Support Devices}\(\}\).

The held-out test set of the CheXpert dataset (518 frontal-view cases) were manually labeled by expert radiologist. This serves as an out-of-distribution (OOD) ground truth benchmark to test generalization across different data distributions, imaging protocols, and labeling schemas. This OOD paradigm is crucial for mimicking real-world clinical deployment, where models must handle variations in patient demographics, equipment, and institutional practices without overfitting to a single source. Metrics included macro-averaged F1-score for multi-label classification. 


\subsubsection{Task and label space.}
To evaluate multi-label prediction, model outputs are taken from the structured \verb|<answer>| block which are already mapped to 14 classes. MedGemini work  considered both multi-select prompt (asking for all labels at once) and per-label Yes/No prompts, and found per-label prompts provided better macro-F1 metric scores. 
In our work we use a single multi-select prompt to be consistent with the overall model workflow. The prompt was "Find abnormalities and support devices", and the reasoning text in \verb|<think>| tags was ignored.

\paragraph{Data splits and preprocessing.}
Training used data only from the MIMIC-CXR training set. The SFT stage used a subset of 100K cases (images + reports), whereas the RL stage used all 227K cases (images + report derived labels). OOD evaluation uses the official CheXpert test set.

\paragraph{Metrics.}
We report macro F$_1$ scores.
Clinical classification tasks are often class-imbalanced: positive (abnormal) cases are relatively rare, yet clinically critical. Metrics that are dominated by the majority class (e.g., accuracy) can therefore be misleading. The F1 score directly balances \emph{precision} and \emph{recall} (sensitivity), making it well-suited to summarize performance when both missed findings and false alarms have tangible clinical costs.

For a given class $k$, with true positives $\mathrm{TP}_k$, false positives $\mathrm{FP}_k$, and false negatives $\mathrm{FN}_k$:
\[
\mathrm{F1}_k=\frac{2\,\mathrm{Precision}_k\cdot \mathrm{Recall}_k}{\mathrm{Precision}_k+\mathrm{Recall}_k}
=\frac{2\,\mathrm{TP}_k}{2\,\mathrm{TP}_k+\mathrm{FP}_k+\mathrm{FN}_k}.
\]
The \textbf{macro F1} averages per-class F1 scores uniformly:
\[
\mathrm{Macro\text{-}F1}=\frac{1}{K}\sum_{k=1}^{K}\mathrm{F1}_k,
\]
giving rare and common conditions equal weight.

\subsubsection{Reporting.}

\begin{table*}[t]
\centering
\caption{Per-class F1 scores \% (higher is better) on CheXpert OOD test set.}
\begin{tabular}{lcccc}
\hline
\textbf{Abnormality class} & \textbf{Ours} & \textbf{MedGemini} & \textbf{Gemini Ultra} & \textbf{MedGemma-4B} \\
\hline
\textbf{Atelectasis}        & 67.2 & 51.1 &  49.7 &  49.0 \\
\textbf{Cardiomegaly}       & 74.7 & 51.5 &  72 &  62.7 \\
\textbf{Consolidation}      & 23.5 & 17.9 & 23 &  31.4 \\
\textbf{Edema}    & 66.7 & 46.2 & 32.7 &  44.3 \\
\textbf{Pleural Effusion}   & 71.1 & 46.6 & 64.4 &  59.5 \\
Pleural Other    & 22.2 & - & - & - \\
Enlarged Cardiomediastinum     & 84.9 & - & - & - \\
Fracture    & 20 & - & - & - \\
Lung Lesion   & 47.1 & - & - & - \\
Lung Opacity    & 88.1 & - & - & - \\
Pneumonia    & 44.4 & - & - & - \\
Pneumothorax     & 57.1 & - & - & - \\
Support Devices     & 93 & - & - & - \\
No Findings    & 79.5 & - & - & - \\

\hline
\textbf{Macro-F1 (5 classes)} & \textbf{60.6} & 48.3 & 42.6 & 48.1  \\
\textbf{Macro-F1 (all)} & 60  & - & - & - \\
\hline
\label{tab:tab_compare1}
\end{tabular}

\vspace{4pt}
\noindent\footnotesize
Notes: The F1 scores results were obtained directly from the corresponding publications. MedGemini and MedGemma reported their performance only for the 5 abnormalities (in bold).
\end{table*}

Table~\ref{tab:tab_compare1} lists per-class F1 scores for the 14 classes of CheXpert test set. MedGemini~\cite{yang2024advancing_medgemini} reported their results only for key 5 abnormalities (macro-F1 48.3), and also reported comparisons to Gemini Ultra~\cite{gemini2023family} model, which we include here for completeness.  MedGemma 4B model~\cite{sellergren2025medgemma} reported their result only for the average macro-F1 score (41.1) for the key 5 abnormalities. We recomputed the individual class F1 scores for the 5 classes using their public model , which resulted in 49.4 average macro-F1 score (consistent with the 48.1 number reported by MedGemma~\cite{sellergren2025medgemma}) . Furthermore the preliminary results for a large MedGemma 20B model lists the macro-F1 score of 49.9.  Our model achieved a strong macro-F1 score for these 5 classes of 60.6, given a similar out-of-distribution testing setup.  

\subsubsection{Ablation studies.}

\begin{table*}
\centering
\caption{Ablation study. Per-class F1 scores (\%) on CheXpert test set.}
\begin{tabular}{lccccc}
\hline
\textbf{Abnormality class} & \textbf{Base} & \textbf{SFT$_{25K}$} & \textbf{GRPO$_{*25K}$}  & \textbf{SFT$_{100K}$} & \textbf{GRPO$_{*100K}$} \\
\hline
\textbf{Atelectasis}    & 35.2 & 42.3 & 65.8 & 47.1 & 67.2 \\
\textbf{Cardiomegaly}       & 39.6 & 57.5 & 73.1 & 60.8 & 74.7  \\
\textbf{Consolidation}      & 0 & 11.3 & 24.8 & 14 & 23.5 \\
\textbf{Edema}    & 36.5 & 42.9 & 65.8 & 60.3 & 66.7  \\
\textbf{Pleural Effusion}   & 23 & 49.6 & 69.4 & 66.1 & 71.1  \\
Pleural Other    & 11.3 & 23.2 & 26.8 & 25 & 22.2  \\
Enl. Cardiomediastinum     & 26 & 42.9 & 83.7 & 55.3 & 84.9  \\
Fracture    & 18.7 & 25.3 & 25.3 & 28.6 & 20  \\
Lung Lesion   & 22.6 & 25.1 & 46.8 & 26.7 & 47.1  \\
Lung Opacity    & 57.1 & 71.6 & 82.6 & 82.2 & 88.1  \\
Pneumonia    & 12.6 & 12.3 & 43.2 & 14.8 & 44.4 \\
Pneumothorax     & 36.7 & 35.9 & 58.4 & 40 & 57.1  \\
Support Devices     & 81.2 & 89.8 & 92.3 & 92.2 & 93  \\
No Findings    & 61.9 & 61.4 & 78.6 & 66.3 & 79.5 \\

\hline
\textbf{Macro-F1 (5 classes)} & 26.9 & 40.72 & 59.8 & 48.5 & 60.6   \\
\textbf{Macro-F1 (all)} & 33 & 42.2 & 59.7 & 49.7 & 60 \\
\hline
\label{tab:tab_ablate1}
\end{tabular}
\end{table*}

We ablate several training configurations and compare them to the base model \emph{Qwen2.5-VL-3B-Instruct}. Because the base VLM does not follow our structured output format, we query it with per-class binary prompts (e.g., ``Is \textit{Atelectasis} present? Answer only Yes or No''). We then evaluate (see Table~\ref{tab:tab_ablate1}):
\begin{itemize}
  \item \textbf{SFT\textsubscript{100K}:} supervised fine-tuning on 100K synthetic reports.
  \item \textbf{GRPO\textsubscript{*100K}:} the final model after GRPO on all 227K cases, initialized from SFT\textsubscript{100K}.
  \item \textbf{SFT\textsubscript{25K}:} low-data SFT on 25K synthetic reports.
  \item \textbf{GRPO\textsubscript{*25K}:} GRPO on all 227K cases initialized from SFT\textsubscript{25K}.
\end{itemize}

Both SFT and GRPO substantially improve over the base model (as expected, since the base model is not medical-specific). GRPO further improves upon SFT despite using only a weak, verifiable reward signal (the abnormality set) without full reasoning text. In the low-data regime (SFT\textsubscript{25K}), accuracy drops relative to SFT\textsubscript{100K}; however, subsequent GRPO on the full 227K set (GRPO\textsubscript{*25K}) nearly recovers to the accuracy of the full model (GRPO\textsubscript{*100K}). This highlights the value of RL (GRPO) for closing the gap when supervised initialization data are limited.

\subsection{Limitations}
\label{limitations}

We acknowledge the following limitations:

\begin{itemize}
    \item \textbf{Hallucinations.} As with most LLM/VLM systems, our model can hallucinate. For example, in one case it incorrectly indicated a pneumothorax; when queried, its “evidence” was the presence of a chest drain tube, likely reflecting spurious co-occurrence in training data. Even in such failures, the model’s explicit reasoning enables auditing and dismissal of erroneous conclusions—an advantage of transparency.
    
    \item \textbf{Single view.} We restrict analysis to a single frontal view (PA or AP) and ignore the lateral view for simplicity. In clinical practice, additional views (and prior studies) are often informative and could improve performance.
    
    \item \textbf{Data size.} Training used only MIMIC-CXR. Some classes were underrepresented (e.g. fracture). Human reasoning annotations were limited; the SFT stage relied primarily on synthetic traces. We expect access to more expert-authored reasoning to further improve the model.
    
    \item \textbf{Indications.} Radiologists routinely use clinical indications (e.g., fever, cough) to guide differential diagnosis. Our model was trained without indications as input, making the task harder. While indications can be provided at inference via follow-up prompts, they were not used during training; principled incorporation of such context should further improve performance.
\end{itemize}

\section{Conclusion}
\label{conclusion}

We introduced a medical reasoning VLM for chest radiography that couples high-fidelity visual encoding with chain-of-thought (CoT) generation, enabling explicit, auditable rationales alongside diagnostic predictions. Our training paradigm blends a reasoning-style SFT warm start with GRPO-based reinforcement learning guided by verifiable, set-level rewards over CheXpert abnormalities, plus formatting and length-aware terms to preserve structured outputs and discourage truncated reasoning. This design teaches the model \emph{how} experts reason—not only \emph{what} they conclude—aligning intermediate steps with observable image evidence and common radiology workflow.

Empirically, the model delivers two complementary benefits. First, in subjective reader studies, experts rated the reasoning traces as accurate, appropriately qualified, and practically useful; full reasoning notably improved trust and confidence and yielded substantial time savings—especially for abnormal studies—when paired with a pre-populated structured report. Second, in numeric OOD evaluation on CheXpert, our approach achieves competitive macro-F1 (and strong per-class F1 on key conditions) compared to general-purpose medical VLMs, while ablations confirm that GRPO further improves over SFT alone even with weak, verifiable supervision. Together, these results indicate that reasoning-first training can enhance both interpretability and performance for chest X-ray analysis.


By releasing NV-Reason-CXR-3B and training code, we aim to catalyze community progress toward trustworthy, explainable medical AI where the quality of reasoning is treated as a first-class objective alongside predictive accuracy. We believe this path—reasoning aligned to verifiable supervision, reported with structured, human-interpretable narratives—offers a practical route to safer adoption, improved auditability, and deeper human–AI collaboration in radiology.

{
  \small
  \bibliographystyle{unsrt}
  \bibliography{paper}
}


\end{document}